\documentclass[pdflatex,sn-mathphys-num]{sn-jnl}

\usepackage{graphicx}%
\usepackage{amsmath,amssymb,amsfonts}%
\usepackage{amsthm}%
\usepackage{mathrsfs}%
\usepackage[title]{appendix}%
\usepackage{xcolor}%
\usepackage{textcomp}%
\usepackage{manyfoot}%
\usepackage{booktabs}\let\cline\cmidrule
\usepackage{pifont}
\newcommand{\cmark}{\ding{51}}%
\newcommand{\xmark}{\ding{55}}%
\usepackage{multirow}

\usepackage{placeins} 

 \usepackage{anyfontsize} 

\usepackage{lineno}

\usepackage{float}
\usepackage{setspace}

\begin{document}

\title[Article Title]{Disease Entity Recognition and Normalization is Improved with Large Language Model Derived Synthetic Normalized Mentions}


\author*[1]{\fnm{Kuleen} \sur{Sasse}}\email{ksasse@uab.edu}

\author[1]{\fnm{Shinjitha} \sur{Vadlakonda}}\email{sgvadlakonda@gmail.com}

\author[1]{\fnm{Richard E.} \sur{Kennedy}}\email{rekenned@uab.edu}

\author*[1]{\fnm{John D.} \sur{Osborne}}\email{ozborn@uab.edu}

\affil[1]{\orgdiv{Department of Medicine}, \orgname{University of Alabama at Birmingham}, \orgaddress{\street{1720 2nd Ave South}, \city{Birmingham}, \postcode{35294}, \state{Alabama}, \country{USA}}}

\abstract{



\textbf{Background:} Machine learning methods for clinical named entity recognition and entity normalization systems can utilize both labeled corpora and Knowledge Graphs (KGs) for learning. However, infrequently occurring concepts may have few mentions in training corpora and lack detailed descriptions or synonyms, even in large KGs. For Disease Entity Recognition (DER) and Disease Entity Normalization (DEN), this can result in fewer high quality training examples relative to the number of known diseases. Large Language Model (LLM) generation of synthetic training examples could improve performance in these information extraction tasks.

\textbf{Methods:} We fine-tuned a LLaMa-2 13B Chat LLM to generate a synthetic corpus containing normalized mentions of concepts from the Unified Medical Language System (UMLS) Disease Semantic Group. We measured overall and Out of Distribution (OOD) performance for DER and DEN, with and without synthetic data augmentation. We evaluated performance on 3 different disease corpora using 4 different data augmentation strategies, assessed using BioBERT for DER and SapBERT and KrissBERT for DEN.

\textbf{Results:} 
Our synthetic data yielded a substantial improvement for DEN, in all 3 training corpora the top 1 accuracy of both SapBERT and KrissBERT improved by 3-9 points in overall performance and by 20-55 points in OOD data. A small improvement (1-2 points) was also seen for DER in overall performance, but only one dataset showed OOD improvement.

\textbf{Conclusion:} 
LLM generation of normalized disease mentions can improve DEN relative to normalization approaches that do not utilize LLMs to augment data with synthetic mentions. Ablation studies indicate that performance gains for DEN were only partially attributable to improvements in OOD performance. The same approach has only a limited ability to improve DER. We make our software and dataset publicly available.

}




\maketitle

\section{Introduction}\label{sec1}

Named Entity Recognition (NER) and Entity Normalization (EN) of those entities to a vocabulary are core tasks in information extraction. NER identifies a mention of an entity in a text span or spans. EN assigns a unique identifier to that mention for a given vocabulary. In the biomedical and clinical space, the NER task is complicated by the diverse lexical forms that biomedical concepts can take, including non-standard names, abbreviations, complex conditions, and misspellings in clinical text \cite{leaman2013dnorm}. Normalization of these entities is also challenging due to the size of vocabularies like SNOMED CT \cite{el2018snomed}(over 360K concepts) or meta-vocabularies like Unified Medical Language System (UMLS)\cite{bodenreider2004unified} with 3.38 million concepts. Moreover, training corpora can suffer from a "long tail" of training entities where the majority of mentioned entities comprise both a small percentage of the total terms and the language that can describe those terms. This phenomena has been documented multiple times. In Fung et al. \cite{fung2010umls}, they analyze counts of normalized entities from the Electronic Health Records (EHRs) of multiple institutions. They find that to get 95\% coverage of all counts, only 21\% of terms had to be used. In Portelli et al. \cite{portelli-etal-2022-generalizing}, all three medical normalization datasets (SMM4H \cite{smm4h-2020-social}, CADEC \cite{karimi2015cadec}, and their proprietary dataset) exhibit long tail issue with MedDRA as the vocabulary as only the top 100-150 terms are represented.

For this work, we focus specifically on Disease Entity Recognition (DER) and Disease Entity Normalization (DEN) which can exhibit both diverse mention descriptions. Furthermore, the wide-ranging frequency of diseases means that the number of mentions of diseases in training datasets often falls off extremely quickly as the disease becomes rare. This falloff contributes to many diseases only having a handful to no mentions while only common diseases receive the majority of the mentions. This leads to a lack of quality and comprehensive training datasets as human annotation is both expensive and time consuming. 

Despite these challenges, a number of datasets have been created to support both DER and DEN. On the biomedical side, the BioCreative 5 CDR corpus \cite{10.1093/database/baw068} and NCBI Disease corpus \cite{DOGAN20141} are the two main disease datasets that utilize PubMed abstracts and papers. These are easily accessible and open source. On the clinical side, the de-identified SemEval 2015 Task 14 \cite{elhadad-etal-2015-semeval} corpus and 2010 i2b2/VA challenge corpus \citep{uzuner20112010} are available and highly cited.
However, these clinical datasets require a data use agreement and careful handling due to the sensitive nature of the data.

An alternative to the generation of human-curated datasets for these tasks is synthetic data to generate synthetic text.
Synthetic data has been explored for a variety of biomedical tasks including Alzheimer's detection \cite{li-etal-2023-two}, creating open source clinical Large Language Models (LLMs) \cite{yang2022large,kweon2024publiclyshareableclinicallarge}, medical dialogue summarization \cite{chintagunta-etal-2021-medically} and medical question answering \cite{guo2023improvingsmalllanguagemodels}. Work focused on using synthetic dataset for clinical NER or EN has been more limited. In Hiebel et al \cite{hiebel-etal-2023-synthetic}, they found that training on synthetic data improved performances of GPT-2 like models to extract spans from clinical text in French. In Meoni et al.\cite{meoni-etal-2023-large}, they focus on using synthetic text to improve models in multilingual clinical domains. In Li et al \cite{li2021synthetic}, they train GPT-2 to generate history and present illness sections of clinical notes which they manually annotate. They then add this data to their NER data to improve their performance. Other work in this space has focused on using synthetic dataset for improving models to find Social and Behavioral Determinants of Health (SBDH). In Mitra et al \cite{mitra2024synthsbdhsyntheticdatasetsocial} and Guevara et al \cite{guevara2024large}, they use "in context learning" \cite{brown2020language} to guide GPT-4 and ChatGPT to generate different SBDH from a set of definitions for each SBDH. 

While other works have focused on NER, using synthetic dataset for EN has not been been explored as thoroughly, despite the widespread use of other data augmentation methods. Munnangi et al \citep{munnangi-etal-2024-fly} did use LLMs for definition augmentation in concept normalization, but did not extend this work to the generation of mentions. Similarly, the REAL system \citep{shlyk2024real} used an LLM for entity normalization, but used RAG \citep{lewis2020retrieval} to link in concepts rather than for data augmentation.
Top performing systems in EN have instead used smaller encoder models instead of LLMS, this includes SapBERT \cite{liu2021selfalignmentpretrainingbiomedicalentity} and CODER \citep{yuan2022coder} which use self-alignment pretraining and leverage UMLS sourced data rather than synthetic data. Similarly, KrissBERT\citep{zhang2022knowledge}, relies on UMLS for data augmentation to support its self supervised training strategy which uses a variety of data generation heuristics to create additional data for contrastive learning achieving State of the Art (SotA) results. Moreover, KrissBERT fully utilizes the context of the system allowing entities to be uniquely identified even when mentions are identical. Finally, data augmentation is used by GenBioEL \citep{yuan-etal-2022-generative} and \citep{Wang2023.12.28.573586} which utilizes knowledge base data augmentation in training a large language model.

In this work we explore the use of LLMs to generated synthetic data for both DER and DEN, and test the utility of this training data improve both DER and DEN under a variety of different training and evaluation scenarios. We evaluate the utility of this data using BioBERT\citep{Lee_2019} for DER and KrissBERT\citep{zhang2022knowledge} and SapBERT\cite{liu2021selfalignmentpretrainingbiomedicalentity} for DEN due to their SotA or near SotA performance, wide-spread use and code availability. 



\section{Methods}

\subsection*{Disease Corpora}
SemEval 2015 Task 14 \cite{elhadad-etal-2015-semeval}, is a clinical text dataset of discharge summaries and radiology reports from the MIMIC II corpus \cite{lee2011open}. This SemEval corpus is annotated with mentions of disorders normalized to the UMLS ontology\cite{bodenreider2004unified}. We use this corpus for the generation of our synthetic data, and again as one of three corpora for downstream evaluation in DER and DEN. This corpora lacks official splits so we used a custom 80/10/10 train/development/test split. This split is used for all tasks.

In addition to SemEval, we use the BC5DR-Disease \cite{10.1093/database/baw068} and the NCBI-Disease \cite{DOGAN20141} for DEN. BC5DR-Disease is the BioCreative 5 Chemical Disease Relation Extraction dataset which consists of PubMed articles annotated with chemical and disease spans that are normalized to the MeSH \cite{lipscomb2000medical} ontology. We take the only the disease spans for our task. NCBI-Disease dataset is a dataset similar to BC5DR which consists of PubMed abstracts annotated with disease spans that are normalized to the MeSH or OMIM \cite{hamosh2005online} ontology. We use the official training and test splits provided for these datasets.

\subsection*{Synthetic Disease Corpora Creation}
LLaMa-2 13B Chat \cite{touvron2023llama2openfoundation} was selected as the LLM to generate synthetic disease mentions, based on its open-source architecture and cost. To build our training dataset, we generated prompt output pairs for each mention in Semeval Task 14 2015. To generate the prompt, we selected the prompt from one of two prompt templates in Figure \ref{tab:prompts-norm}, depending on whether the mention had a definition in UMLS \cite{bodenreider2004unified}. We use only UMLS concepts within the Disorders Semantic Group \cite{bodenreider2004unified} to target the disease for our model to generate.
Additionally, we give the prompt the UMLS alternative preferred English name and the first UMLS provided definition if one is available. For each UMLS disease concept, we generated 5 different versions of the mention to increase the chance of a generating a correctly formatted enclosing \textless1CUI\textgreater tags and to provide diversity in training examples. These tags allow the disease entity to be associated with the corresponding spans of the generated text.
To create the corresponding output, we include the mention-containing sentence as well as the two sentences before and after that sentence. The goal is to provide the model as much relevant context about the disease as possible, leveraging the text generation capabilities of the LLM.

\begin{table}[ht]
    \centering
    \begin{tabular}{p{0.23\linewidth} | p{0.7\linewidth}}

        \textbf{UMLS Definition?} & \textbf{Prompt}\\
        \midrule[.1em]
        Yes & Pretend you are a physician: Write a clinical note for a patient that mentions the condition MENTION either explicitly or as a synonym or abbreviation to this condition. It is also known as LIST OF NAMES OF MENTION IN UMLS. It is defined as UMLS DEFINITION. Place tokens \textless1CUI\textgreater before and after the mention of this condition. For example \textless1CUI\textgreater MENTION \textless 1CUI\textgreater.\\
        \hline
        No & Pretend you are a physician: Write a clinical note for a patient that mentions the condition MENTION either explicitly or as a synonym or abbreviation to this condition. It is also known as LIST OF NAMES OF MENTION IN UMLS. For example \textless1CUI\textgreater MENTION \textless1CUI\textgreater.\\
        \hline
    \end{tabular}
    \vspace*{0mm}
    \caption{Prompts Used in Training LLaMa-2 13B Chat Model. The arbitrarily chosen delimiter \textless1CUI\textgreater is used to enclose the mention and identify the text associated with the CUI.}
    \label{tab:prompts-norm}
\end{table}

We supervised fine-tuned with cross entropy loss our LLaMa-2 13B Chat \cite{touvron2023llama2openfoundation} model on our SemEval Task 14 2015 training dataset  using Quantized Low Rank Adaptation (Q-LoRa) \cite{dettmers2023qloraefficientfinetuningquantized}. We trained our model at 4 bit precision for 2 epochs with a learning rate of 2e-5 with a rank of 64 for our LoRa \cite{hu2021loralowrankadaptationlarge} adapter.

\begin{figure}
    \centering
    \includegraphics[width=\linewidth]{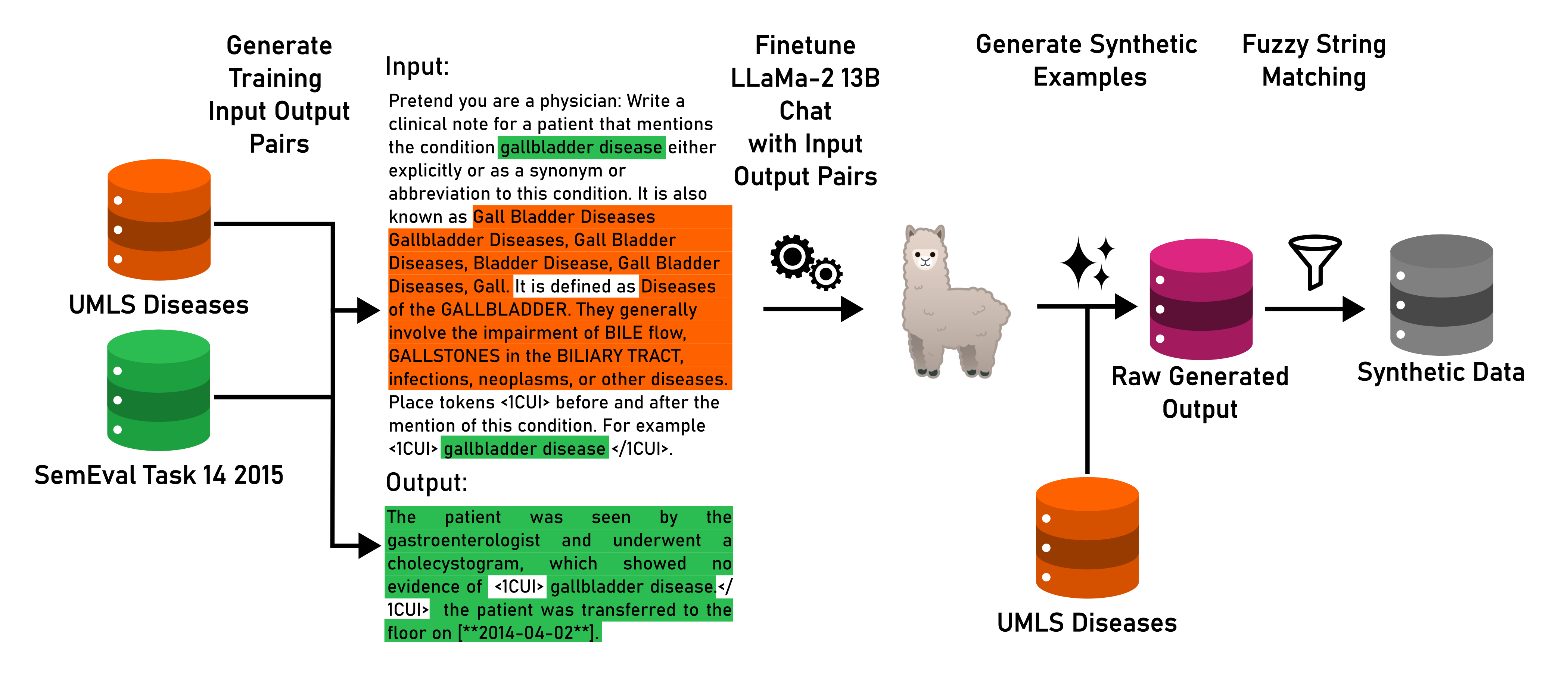}
    \caption{An illustration of the synthetic data generation process.}
    \label{fig:DER}
\end{figure}

To get the initial labels for each synthetic disease mention, we used fuzzy string matching from the Python regex package to extract the disease name from the generated mention. Under the hood, the package adds an edit distance feature to a typical regex package. We searched the synthetic disease mention for the disease mention used in the prompt to generate it. We removed any note where a disease mention could not be identified by fuzzy matching with a budget of 4 edits. After this step, we were left with around 128,000 mentions for 47654 CUIs out of 53432 Disorder Group CUIs available. For each CUI, an average of 3 unique mentions were generated. All notes were unique and were not copies of the training data.

\begin{table}[ht]
    \centering
    \begin{tabular}{p{0.23\linewidth} | p{0.7\linewidth}}
        \textbf{Disease} & \textbf{Output}\\
        \midrule[.1em]
         Beryllium Disease & Discharge date:  [**2013-08-24**]  date of birth:   [**1947-12-28**]  sex:  m  service:  cardiothoracic  allergies:  patient recorded as having no known allergies to drugs  attending:[**attending info 212**]  chief complaint:  chest pain  major surgical or invasive procedure:   \textless 1CUI\textgreater beryllium dis \textless/1CUI\textgreater   history of present illness:  mr. [**lastname 707**] is a 65-year-old male who presented to the [**hospital1 199**] [**doctor first name 3**] with chest pain.  he is a former worker at a beryllium plant and was exposed to beryllium.\\
        \hline
        Metastatic Pancreatic Neuroendocrine Tumor & Discharge date:   [**2018-11-03**]  date of birth:  [**1934-12-02**]             sex:   m  service: medicine  allergies:  patient recorded as having no known allergies to drugs  attending:[**attending info 1699**]  chief complaint: h/o  \textless 1CUI\textgreater metastatic pancreatic neuroendocrine tumor \textless/1CUI\textgreater .  major surgical or invasive procedure: 1.  resection of metastatic pancreatic neuroendocrine tumor.\\ 
        \hline
    \end{tabular}
    \vspace*{-1mm}
    \caption{Examples of Generated Disease Mentions (GDMs)}
    \label{tab:examples_of_generated}
\end{table}

\subsection{Data Augmentation Strategies}
We apply four different data augmentation strategies to fine-tune our DER and DEN models, based on the overlap between the generated concepts present in the synthetic corpus and whether those same concepts are found in the training and/or test splits of the evaluation corpora. Our first method, ``\textit{\underline{Naive}}'' adds \textit{ALL the synthetic data}. Our second method, ``\textit{\underline{Ideal}}'' adds the portion of the synthetic dataset \textit{appearing in the test split}. The third method adds only the synthetic examples \textit{NOT in the train set}, even when they are also found in the test split. We term this strategy  ``\textit{\underline{Supplemental}}'' and it represents a preference to using synthetic data only when human curated data is not available.
\begin{table}[ht]
\addtolength{\tabcolsep}{-0.22em}
\begin{tabular}{llrr|rrrrrrrr}
\textbf{}                         & \textbf{}                                & \multicolumn{2}{c}{\textbf{Original}}      & \multicolumn{2}{c}{\textbf{Naive}}                                     & \multicolumn{2}{c}{\textbf{Ideal}}                                     & \multicolumn{2}{c}{\textbf{Supp.}}                                     & \multicolumn{2}{c}{\textbf{Ablation}} \\
\cmidrule(lr){3-4}\cmidrule(lr){5-12}
\textbf{Dataset}                 & \multicolumn{1}{l|}{\textbf{Split}} & \multicolumn{1}{l}{\textbf{CUI}} & \multicolumn{1}{l}{\textbf{Mnt.}} & \multicolumn{1}{l}{\textbf{CUI}} & \multicolumn{1}{l}{\textbf{Mnt.}} & \multicolumn{1}{l}{\textbf{CUI}} & \multicolumn{1}{l}{\textbf{Mnt.}} & \multicolumn{1}{l}{\textbf{CUI}} & \multicolumn{1}{l}{\textbf{Mnt.}} & \multicolumn{1}{l}{\textbf{CUI}} & \multicolumn{1}{l}{\textbf{Mnt.}}                           \\ 
\toprule
\multirow{2}{*}{\textbf{SemEval}} & \multicolumn{1}{l|}{Train}          &1689& 16220                              & 920                               & 128914                             & 212                               & 749                                & \xmark                                 & 126243                             & 708                               & 128165 \\
                                  & \multicolumn{1}{l|}{Test}           & 383                               & 1523                               & 250                               & 1523                               & 250                               & 1523                               & 38                                & 1523                               & \xmark                                 & 1523  \\ 
\cline{1-12}
\multirow{2}{*}{\textbf{BC5DR}}   & \multicolumn{1}{l|}{Train}          & 634                               & 4318                               & 398                               & 128914                             & 94                                & 1255                               & \xmark                                & 127628                             & 304                               & 127658 \\
                                  & \multicolumn{1}{l|}{Test}           & 196                               & 4135                               & 133                               & 4135                               & 133                               & 4135                               & 39                                & 4135                               & \xmark                                 & 4135  \\ \cline{1-12}
\multirow{2}{*}{\textbf{NCBI}}    & \multicolumn{1}{l|}{Train}          & 655                               & 5091                               & 435                               & 128914                             & 256                               & 384                                & \xmark                                 & 127721                             & 179                               & 128528  \\
                                  & \multicolumn{1}{l|}{Test}           & 639                               & 952                                & 428                               & 952                                & 428                               & 952                                & 172                               & 952                                & \xmark                                 & 952 \\ \cline{1-12}
\end{tabular}%
\caption{The total number of disease concepts (CUIs) and mentions for the original dataset are shown under the Original heading. For evaluation of each augmentation strategy, the original held-out test split is used. For training, the total number of mentions (Mnt.) used is shown. The number of concepts (CUIs) from generated mentions that overlap concepts in the the original train and test for that dataset are shown in the CUI columns for each strategy.}
\label{tab:DataAugmentationStrategies}
\end{table}
The fourth method and final adds only the synthetic examples \textit{not in the test set} named ``\textit{\underline{Ablation}}''. An overview of the original and augmented datasets is presented in Table \ref{tab:DataAugmentationStrategies}.

\subsection*{Evaluation of Synthetic Dataset on Disease Entity Recognition}
We frame this problem as a disorder span prediction problem using a binary tagging framework where either a token is a part of a disease mention or not. We used BioBERT-base \cite{Lee_2019} as a baseline model, a BERT model that has been pretrained on a large corpus of biomedical text. We chose an encoder only model over larger decoder only models as decoder only models have often struggled on Biomedical NER tasks in comparison to smaller encoder models \cite{munnangi-etal-2024-fly}. We trained this model on the train split of the dataset for 3 epochs with early stopping based on loss on the validation set.

After training the baseline models, we use these models to label our synthetic data in order to label spans of diseases mentioned outside the fuzzy matched span. We call this our \underline{\textit{Labelled}} version of the dataset. 

For our synthetic text models, we finetuned BioBERT-base on either the fuzzy matched or \textit{Labelled} synthetic text for 10 epochs with early stopping on the gold validation set. We then further finetuned that synthetic model on the gold training set for 3 epochs with early stopping based on loss on the gold validation set.

\begin{figure}
    \centering
    \includegraphics[width=\linewidth]{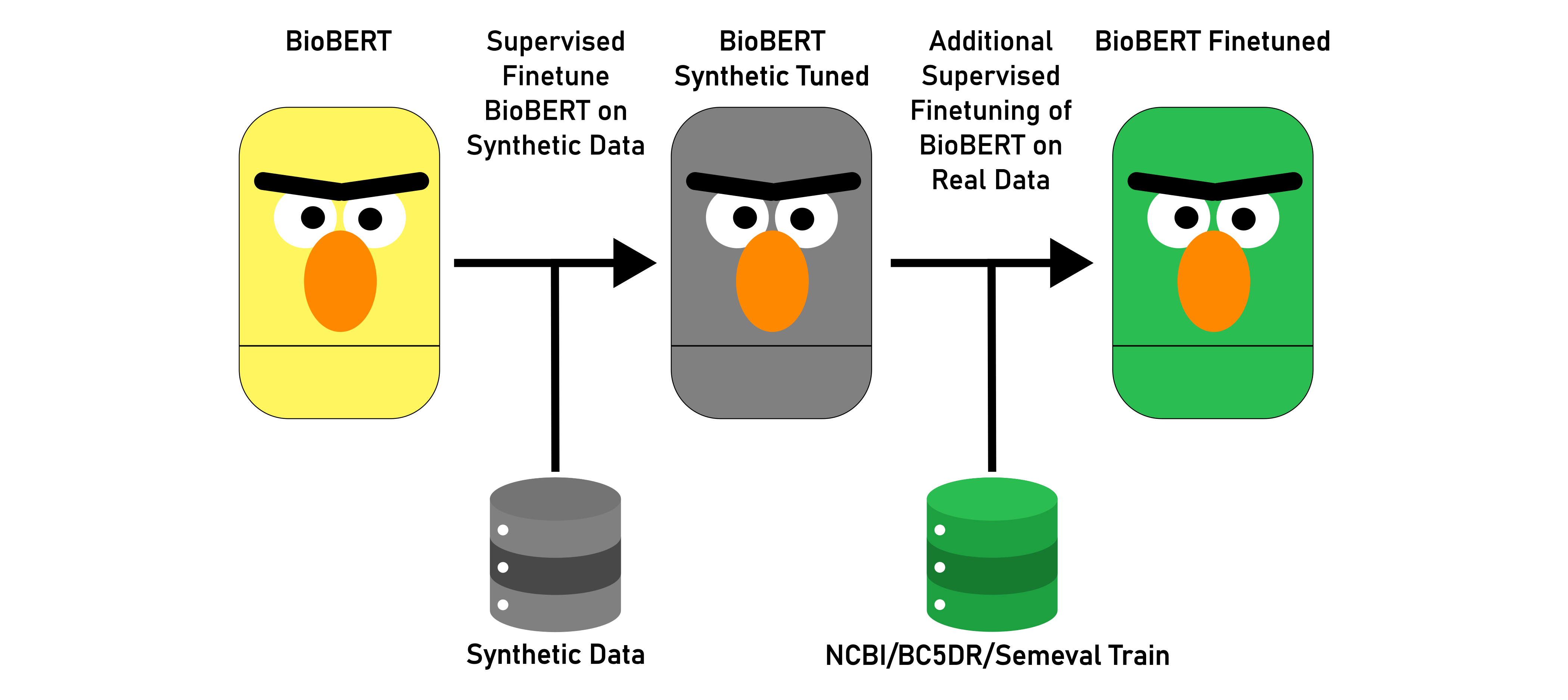}
    \caption{An illustration of the DER Training Process. BioBERT images were adapted from Jay Alammar's Illustrated BERT\cite{jalammarIllustratedBERT} and permission to include in this work is granted by the CC BY-NC-SA 4.0 license\citep{licensesattribution}.}
    \label{fig:DER_Train}
\end{figure}

To evaluate these methods, we measured the overall Precision, Recall, and F1 on the test set using the seqeval \cite{seqeval} library. In addition to overall results, we measured the Out Of Distribution (OOD) performance using the same metrics. OOD performance is the performance only on mentions of disorders that do not appear in the training dataset. We select this subset using the CUIs provided by the dataset. We ran our experiments 5 times with different random seeds and averaged their scores. 

\subsection*{Evaluation of Synthetic Dataset on Disease Entity Normalization}
To evaluate our synthetic mentions for Disease Entity Normalization, we used three datasets described in Table \ref{tab:DataAugmentationStrategies} adapted to a normalization task. As our synthetic data use UMLS normalization identifiers, we mapped MeSH and OMIM identifiers from the BC5DR and NCBI datasets to UMLS identifiers. 
We evaluate our synthetic dataset on two neural models and include two other text based methods as baselines. For our neural models, we chose the widely used KrissBERT \cite{zhang2022knowledgerichselfsupervisionbiomedicalentity} and SapBERT \cite{liu2021selfalignmentpretrainingbiomedicalentity} models which have SotA performance or close to SotA performance. Both models were adapted to use our synthetic corpus. SapBERT is a model that was contrastively pretrained on mention pairs sampled from UMLS. To adapt SapBERT model to our synthetic dataset, all mentions are passed through the model and a space of vectors are constructed. To run inference, the test span is passed through the model and the nearest vector to the output vector is taken as the label. 

\begin{figure}
    \centering
    \includegraphics[width=\linewidth]{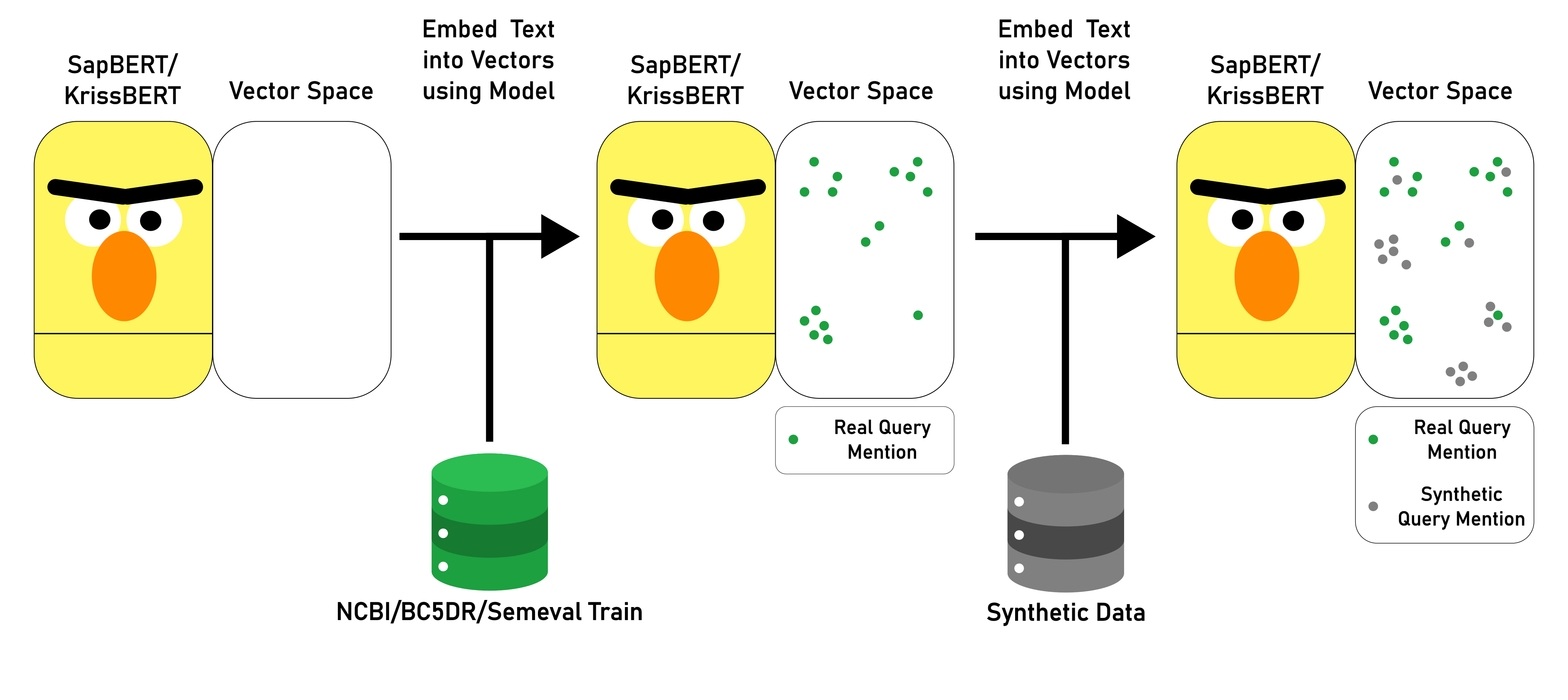}
    \caption{An illustration of the DEN Training Process. BioBERT images were adapted from Jay Alammar's Illustrated BERT\cite{jalammarIllustratedBERT} and permission to include in this work is granted by the CC BY-NC-SA 4.0 license\citep{licensesattribution}. }
    \label{fig:DEN}
\end{figure}

KrissBERT\cite{zhang2022knowledge} is a model that was contrastively pretrained on full-text mentions from PubMed, using UMLS string matching to identify concepts and enhancing performance with mention masking and replacement. To adapt the model to a dataset, all the full-text mentions are passed through the model to create a space of all the vectors of the mentions. To run inference on KrissBERT, the test mentioned is fed into KrissBERT and the $k$th closest vectors to its corresponding output vector are found using a nearest neighbor lookup. The label of the output vector is decided by majority vote of those nearest neighbors.  

For our text based baselines, we chose QuickUMLS \cite{soldaini2016quickumls} and SciSpacy \cite{neumann-etal-2019-scispacy}. QuickUMLS is string matching based method that relies on knowledge of forms in order to determine similarity and uses overlap of tokens. SciSpacy is similar as to QuickUMLS as it is a string matching based method but it uses 3-gram character overlap instead of tokens. As these text models cannot be trained, we do not add the synthetic dataset. For our neural models, we apply all four of our data augmentation strategies, \textit{Naive}, \textit{Ideal}, \textit{Supplemental}, \textit{Ablation} and show the baseline performance of both SapBERT\citep{liu2021selfalignmentpretrainingbiomedicalentity} and KrissBERT\citep{zhang2021knowledge} without our data augmentation.

To measure the performance of the models, we measure both Accuracy @ $k$ and Out of Distribution Accuracy @ $k$ for $k = [1,5,50]$. To measure Accuracy @ $k$, we measure whether the correct CUI appears the top $k$ values returned from the model similarly to Garda et al.\cite{garda2023belb}. Out of Distribution Accuracy similar to Accuracy but its only calculated on mentions that are not in the training dataset.

\subsection*{Statistical Significance}
To evaluate statistical significance for DER performance, we used Mann Whitney U-Tests to compare each strategy to the baseline on Precision, Recall, F1 and Accuracy scores, similar to Guevara et al. \cite{guevara2024large}. We use a threshold of $\alpha = 0.05$. For DEN evaluation we run our experiments only once, as the entire process is deterministic. The vectors are calculated for each text by the same model without dropout and the nearest neighbor searches are deterministic as they are exact. 

\section{Results}\label{sec2}
 
\subsection*{Disease Entity Recognition (DER) Results}
A summary of DER results for all \textit{Labelled} datasets and strategies is shown in Table \ref{tab:All_DER}, including both overall and OOD results. Overall, results without the baseline model \textit{Labelling} perform below those with that step and there are no significant results above the baseline model without the baseline model labelling step. See Appendix \ref{secB1} for a side by side comparison of results with and without the \textit{Labelling} step. 

Overall, there are small but statistically significantly gains in overall performance using both the \textit{Naive} strategy of adding all generated mentions and the \textit{Supplemental} strategy of adding only Generated Disease Mentions (GDMs) whose concepts (CUIs) were not already represented in the original training splits. This performance improvement is seen across all 3 datasets, with slightly higher performance on the BC5CDR and NCBI datasets. 
The \textit{Ideal} strategy which uses only GDMs present in the original test data split showed no significant difference from the baseline model, suggesting overall improvements in performance are not simply the result of adding in test concept GDMs. This is further supported by the \textit{Ablation} strategy results where GDMs with CUIs that overlap the test data split CUIs are removed from augmented training, even if those CUIs were already present in the training dataset. The \textit{Ablation} strategy outperformed even the \textit{Naive} strategy on the BC5CDR and NCBI dataset. However its performance was not significantly better on the SemEval dataset. 

\begin{table}[htbp]
\resizebox{\textwidth}{!}{%
\begin{tabular}{ l l | r r r | r r r }
& & \multicolumn{3}{c}{\textbf{Overall}}  & \multicolumn{3}{c}{\textbf{Out of Distribution}} \\ 
\textbf{Dataset} & \textbf{Augmentation} & 
\textbf{Prec} & \textbf{Rec} & \textbf{F1} & 
\textbf{Prec} & \textbf{Rec} & \textbf{F1} \\ 
\toprule

\multirow{5}{*}{SemEval}  
& None (Baseline)
  & 0.7915 & 0.8000 & 0.7957 & 0.5127 & 0.7863 & 0.6190 \\
& Naive
 & \textbf{0.7974} & \underline{\textbf{0.8314}} & \underline{\textbf{0.8140}} & \textbf{0.5354} & 0.7710 & \underline{\textbf{0.6316}} \\
& Ideal
 & \textbf{0.8033} & 0.7839 & 0.7933 & 0.4564 & \textbf{0.8107} & 0.5829 \\
& Supplemental
 & \textbf{0.7989} & \underline{\textbf{0.8149}} & \underline{\textbf{0.8068}} & 0.4892 & 0.7748 & 0.5996 \\
& Ablation
 & \textbf{0.7972} & \textbf{0.8174} & \textbf{0.8072} & 0.4948 & \textbf{0.7878} & 0.6076 \\
\hline
 
\multirow{5}{*}{BC5CDR}  
& None (Baseline)
  & 0.8229 & 0.7578 & 0.7888 & 0.4908 & 0.8453 & 0.6200 \\
& Naive
 & \underline{\textbf{0.8373}} & \underline{\textbf{0.7911}} & \underline{\textbf{0.8135}} & 0.4843 & \textbf{0.8498} & 0.6168 \\
& Ideal
 & \textbf{0.8347} & \textbf{0.7627} & \textbf{0.7979} & 0.4617 & \textbf{0.8659} & 0.6018 \\
& Supplemental
 & \underline{\textbf{0.8411}} & \underline{\textbf{0.7933}} & \underline{\textbf{0.8164}} & 0.4770 & \textbf{0.8495} & 0.6105 \\
& Ablation
 & \underline{\textbf{0.8434}} & \underline{\textbf{0.7943}} & \underline{\textbf{0.8181}} & 0.4832 & \textbf{0.8512} & 0.6163 \\
\hline

\multirow{5}{*}{NCBI}  
& None (Baseline)
  & 0.8587 & 0.8052 & 0.8310 & 0.6648 & 0.8932 & 0.7622 \\
& Naive
 & \underline{\textbf{0.6348}} & \underline{\textbf{0.8404}} & \underline{\textbf{0.8548}} & 0.6614 & 0.8773 & 0.7539 \\
& Ideal
 & 0.8547 & 0.7905 & 0.8213 & 0.6308 & 0.8827 & 0.7347 \\
& Supplemental
 & \underline{\textbf{0.8614}} & \underline{\textbf{0.8334}} & \underline{\textbf{0.8471}} & 0.6532 & 0.8789 & 0.7539 \\
& Ablation
 & \underline{\textbf{0.8738}} & \underline{\textbf{0.8415}} & \underline{\textbf{0.8573}} & 0.6633 & 0.8910 & 0.7604 \\
\hline

\end{tabular}%
}
\vspace*{0mm}
\caption{Overall and Out of Distribution (OOD) DER Results. \textbf{Bold} is a model that beats the baseline, \underline{underline} means the difference was statistically significant based on a Mann-Whitney U test with $\alpha = 0.05$.}
\label{tab:All_DER}
\end{table}

In contrast to overall performance, in the OOD evaluation only the \textit{Naive} strategy generated a small, but statistically significant improvement in F1 score and only for the SemEval dataset. Results are shown in the last 3 columns of Table \ref{tab:All_DER}. In general, OOD performance differs little from the baseline, although there is an overall trend to slightly higher recall at the expense of precision in BC5CDR.



\subsection*{Disease Entity Normalization Results}

In contrast to the DER results, the effect of adding synthetic data to overall performance is both larger and broader reaching. Across all 3 datasets and normalization methods we see significant improvements to both overall and OOD DEN performance. The simple \textit{Naive} strategy of adding all the synthetic data generated works consistently across every metric and dataset. Unlike DER, the \textit{Ablation} strategy performs relatively poorly in comparison to the \textit{Ideal} strategy indicating that much of the benefit for DEN comes for the generation of OOD concepts and mentions. However, given the generally consistently higher performance of \textit{Naive} to \textit{Ideal} in overall performance (1-3 points), as well as higher performance across a range of OOD accuracy, we can infer that higher performance is not just the result of the presence of GDM concepts in the test splits.


\begin{table}[ht]
\resizebox{\textwidth}{!}{%
\begin{tabular}{ l l | r r r | r r r }
& & \multicolumn{3}{c}{\textbf{Accuracy}}  & \multicolumn{3}{c}{\textbf{OOD Accuracy}} \\ 
\textbf{Model} & \textbf{Augmentation} & \textbf{Top 1} & \textbf{Top 5} & \textbf{Top 50} & \textbf{Top 1} & \textbf{Top 5} & \textbf{Top 50} \\ \toprule
SciSpacy  & N/A        & 0.5257                         & N/A      & N/A        & N/A        & N/A          & N/A            \\ 
QuickUMLS & N/A        & 0.3630                         & N/A      & N/A        & N/A        & N/A          & N/A           \\ \midrule
\multirow{5}{*}{KrissBERT}  & None (Baseline)   & 0.7604                          & 0.8502                           & 0.8705                            & 0.0000                             & 0.0000                              & 0.0000                               \\  
 & \textit{Naive}     & \textbf{0.7903}                         & \textbf{0.8907}                          & \textbf{0.9225}                            & \textbf{0.2199}                             & \textbf{0.3191}                             & \textbf{0.4043}                               \\ 
 & Ideal   & 0.7542                         & \textbf{0.8722}& \textbf{0.9137}& \textbf{0.1702}& \textbf{0.2695}& \textbf{0.3617}\\ 
 & Supplemental & 0.7498                         & \textbf{0.8678}                           & \textbf{0.8934}                           & \textbf{0.1702}                             & \textbf{0.2695}                              & \textbf{0.3617}                                \\ 
  & Ablation & 0.7260                         & 0.8326                           & 0.8678                           & 0.0000                             & 0.0000                              & 0.0000                                \\ \midrule
\multirow{5}{*}{SapBERT}    & None (Baseline)   & 0.7344                         & 0.8119                           & 0.8599                            & 0.0000                             & 0.0000                              & 0.0000                               \\ 
  & \textit{Naive}     & \textbf{0.7756 }                        & \textbf{0.8586}                          & \textbf{0.9075}                           & \textbf{0.2862}                             & \textbf{0.3569}                            & \textbf{0.3755}                               \\ 
 & Ideal   & 0.7366                          & 0.8409                           & 0.8948                            & 0.2082                              & 0.2714                               & 0.3457                                \\ 
 & \textit{Supplemental} & \textbf{0.7493} & \textbf{0.8404} & \textbf{0.8939} & \textbf{0.1896} & \textbf{0.2714} & \textbf{0.3457}\\ 
  & Ablation &0.7131& 0.8033& 0.8518& 0.0000& 0.0000& 0.0000\\\hline
\end{tabular}%
}
\vspace*{0mm}
\caption{SemEval 2015 Task 14 Synthetic Normalization Results. Accuracy (Acc) is equivalent to Recall. \textbf{Bold} is a model that beats the baseline, \textit{italics} means the model beat the baseline at all thresholds. Our baseline results differ from previously published results due to our requirement for the normalized identifier to be unique and the lack of official splits for this dataset.}
\label{tab:Semeval}
\end{table}


\begin{table}[htbp]
\resizebox{\textwidth}{!}{%
\begin{tabular}{ l l | r r r | r r r }
& & \multicolumn{3}{c}{\textbf{Accuracy}}  & \multicolumn{3}{c}{\textbf{OOD Accuracy}} \\ 
\textbf{Model} & \textbf{Augmentation} & \textbf{Top 1} & \textbf{Top 5} & \textbf{Top 50} & \textbf{Top 1} & \textbf{Top 5} & \textbf{Top 50} \\ \toprule
 
SciSpacy  & N/A          & 0.3986                                              & N/A                                                  & N/A                                                   & N/A                                                     & N/A                                                      & N/A                                                       \\
QuickUMLS & N/A          & 0.2703                                              & N/A                                                  & N/A                                                   & N/A                                                     & N/A                                                      & N/A                                                       \\\midrule
\multirow{5}{*}{KrissBERT} & None (Baseline)     & 0.7677& 0.8263& 0.8491& 0.0000                                                       & 0.0000                                                        & 0.0000                                                         \\
 & \textit{Naive}        & \textbf{0.8314}& \textbf{0.9242}& \textbf{0.9531}& \textbf{0.4240}& \textbf{0.6581}& \textbf{0.7128}\\
 & Ideal        & 0.7203& \textbf{0.8659}& \textbf{0.9262}& \textbf{0.2295}& \textbf{0.4514}& \textbf{0.6459}\\
 & Supplemental & 0.7112& \textbf{0.8614}& \textbf{0.9249}& \textbf{0.2340}& \textbf{0.4544}& \textbf{0.6474}\\
 & Ablation & 0.6765& 0.7945& 0.8294 & 0.0000& 0.0000& 0.0000\\\midrule
\multirow{5}{*}{SapBERT}   & None (Baseline)     & 0.7713& 0.7927& 0.8297& 0.0000& 0.0000& 0.0000
\\
   & \textit{Naive}& \textbf{0.8568}& \textbf{0.8866}& \textbf{0.9323}& \textbf{0.5500}& \textbf{0.6112}& \textbf{0.6860}\\
   & Ideal        & 0.7033& \textbf{0.8230}& \textbf{0.9006}& \textbf{0.3667}& \textbf{0.4736}& \textbf{0.6173}\\
   & Supplemental & 0.6959& \textbf{0.8154}& \textbf{0.8943}& \textbf{0.3659}& \textbf{0.4736}& \textbf{0.6173}\\
   & Ablation & 0.6595& 0.7260& 0.7746 & 0.0000& 0.0000& 0.0000\\\hline
    \end{tabular}
    }
    \vspace*{0mm}
    \caption{BC5DR Synthetic Normalization Results. \textbf{Bold} is a model that beats the baseline, \textit{italics} means the model beat the baseline at all thresholds.}
\label{tab:BC5DR-Disease}
\end{table}


\begin{table}[htbp]
\resizebox{\textwidth}{!}{%
\begin{tabular}{ l l | r r r | r r r }
& & \multicolumn{3}{c}{\textbf{Accuracy}}  & \multicolumn{3}{c}{\textbf{OOD Accuracy}} \\ 
\textbf{Model} & \textbf{Augmentation} & \textbf{Top 1} & \textbf{Top 5} & \textbf{Top 50} & \textbf{Top 1} & \textbf{Top 5} & \textbf{Top 50} \\ \toprule

SciSpacy  & N/A          & 0.3986                                              & N/A                                                  & N/A                                                   & N/A                                                     & N/A                                                      & N/A                                                       \\ 
QuickUMLS & N/A          & 0.2703                                              & N/A                                                  & N/A                                                   & N/A                                                     & N/A                                                      & N/A                                                       \\ \midrule
\multirow{5}{*}{KrissBERT} & None (Baseline)     & 0.7604& 0.8502& 0.8705& 0.0000& 0.0000& 0.0000
\\
 & \textit{Naive}        & \textbf{0.7903}& \textbf{0.8907}& \textbf{0.9225}& \textbf{0.2199}& \textbf{0.3191}& \textbf{0.4043}
\\
 & Ideal        & 0.7542& \textbf{0.8722}& \textbf{0.9137}& \textbf{0.1702}&\textbf{0.2695}& \textbf{0.3617}
\\
 & Supplemental & 0.7498& \textbf{0.8678}& \textbf{0.9137}& \textbf{0.1702}& \textbf{0.2695}& \textbf{0.3617}\\
 & Ablation & 0.7017& 0.7651& 0.8039& 0.0000& 0.0000& 0.0000 \\ \midrule
\multirow{5}{*}{SapBERT}   & None (Baseline)     & 0.7276& 0.7582& 0.7862& 0.0000& 0.0000& 0.0000
\\
   & Naive        & \textbf{0.7730}&\textbf{0.8152}& \textbf{0.8622}& \textbf{0.2398}& \textbf{0.2982}& \textbf{0.4211}
\\
   & Ideal        & 0.7080& \textbf{0.7693}& \textbf{0.8379}& \textbf{0.1637}& \textbf{0.1988}& \textbf{0.3392}
\\
   & Supplemental & 0.6954& \textbf{0.7624}& \textbf{0.8358}& \textbf{0.1637}& \textbf{0.1988}& \textbf{0.3392}\\
   & Ablation & 0.6479& 	0.7469& \textbf{0.8035}& 0.0000& 0.0000& 0.0000\\\hline        
\end{tabular}%
}
\vspace*{0mm}
\caption{NCBI-Disease Synthetic Normalization Results. \textbf{Bold} is a model that beats the baseline, \textit{italics} means the model beat the baseline at all thresholds.}
\label{tab:NCBI-Disease}
\end{table}








\section{Discussion}\label{sec12}
Our results indicate that LLM GDMs can improve overall and OOD performance in both Named Entity Recognition (NER) and Entity Normalization (EN) when those entities are diseases. While we restricted our studies to disease due to the availability of public datasets, in principle our DER and DEN approach should work for other entity types. One important advantage of our approach in comparison to Li et al \cite{li2021synthetic} is that we do not rely a manual annotation step. We generate our annotations with our synthetic text and our method can be easily extended to multiple entities and languages. We think the key to the success of our method versus approaches that either use LLMs to generate knowledge graph sourced definitions only\citep{munnangi-etal-2024-fly}, definition and synonyms without mentions\citep{yuan2022generative} or use self-supervision \citep{zhang2021knowledge} is in part due to the limitations of existing knowledge resources like UMLS. As shown in Appendix \ref{tab:cuiratios}, nearly a third of UMLS disease concepts lack synonyms and most lack formal definitions. Prompting LLMs to generate specific mentions of concepts with known offsets allows LLMs to leverage their large pool of pretraining data to create realistic looking mentions for training. Therefore unlike \cite{meoni-etal-2023-large}, we use our large language model to generate synthetic "notes" (GDMs) directly instead of trying to use it for weak supervision. 

Other benefits of our synthetic augmentation approach include avoiding inadvertent PHI release that can occur using real data. Moreover, this approach has the ability to combine improvements in LLM text generation abilities alongside NER and EN algorithm improvements. Our resulting synthetic mentions can be also interpreted by humans, and therefore can refined or cleaned by humans as well as software. This knowledge transfer is preferred over distillation approaches (where knowledge is transferred from a larger general-purpose LLMs) that rely on model weight adjustments that are not subject to easy interpretation.

Our results indicate that DER benefits less than DEN from synthetic augmented data. This may be due to the greater sparsity of exemplar classes in large vocabularies. DER needs only recognize mentions at a disease level so the variety of new synthetic language to describe disease may be less significant than new entity example classes from a vocabulary. 
Additionally, the lack of considerable gains in OOD performance for DER models on BC5DR and NCBI may be due to domain shift, since our generations are trained on SemEval, a clinical domain dataset. 
Finally, the DEN models evaluated require relatively few examples to get the right answer, as adding synthetic examples fills in their dense vector space whereas DER models have to overcome their pretraining in order to detect these examples. In terms of OOD performance, the normalization models have the most difficult time with OOD items because if the term is not provided to them in the training data, it will never be chosen. Therefore, adding any form of synthetic data is good because it increases the probability of a known concept at inference time.

Our study had a few limitations, including its reliance for generation on a single open-source LLM, LLaMa-2 13B Chat which is aligned for instruction following. We initially tried smaller models from the T5 family\citep{2020t5} but these smaller, older models generated no statistically significant performance improvements and we suspect additional performance improvements are possible with more recent and larger commercial models. However, our preference is to conduct reproducible research on open models, which also allows us to better control and understand the fine-tuning process. Future work will examine the next generation of LLaMa models.

\section{Conclusion}\label{sec13}

We demonstrate that LLMs can be generate synthetic data for both DER and DEN and make this data publicly available. We show that this data can generate minor performance improvements to DER and larger performance improvements in DEN. Furthermore, our results suggest that the improvement to DEN is not just the result of greater test split concept coverage or OOD performance, but that the incorporation of additional mentions creates better vector space representations for normalization. We make our software publicly available.


\backmatter


\section*{Availability of data and materials}
The dataset and code supporting the conclusions of this article are available in the synth-der-den repository, 
\texttt{https://github.com/KuleenS/synth-der-den}.

\section*{Abbreviations}
\begin{itemize}
    \item[] {\bf Avg:} Average
    \item[] {\bf Acc:} Accuracy
    \item[] {\bf BC5DR:} BioCreative 5 Chemical Disease Relation Extraction
    \item[] {\bf BERT:} Bidirectional Encoder Representations from Transformers
    \item[] {\bf BioBERT} Biomedical BERT
    \item[] {\bf CUI} Concept Unique Identifier
    \item[] {\bf DEN:} Disease Entity Normalization
    \item[] {\bf DER:} Disease Entity Recognition
    \item[] {\bf EHR:} Electronic Health Record
    \item[] {\bf EN:} Entity Normalization
    \item[] {\bf ER:} Entity Recognition
    \item[] {\bf F1:} F1 Score
    \item[] {\bf GDM:} Generated Disease Mention
    \item[] {\bf GPT:} Generative Pre-trained Transformer 
    \item[] {\bf HPC:} High Performance Computing
    \item[] {\bf LLM:} Large Language Model
    \item[] {\bf MeSH:} Medical Subject Headings
    \item[] {\bf MIMIC:} Medical Information Mart for Intensive Care 
    \item[] {\bf Mnts.:} Mention
    \item[] {\bf NA:} Not Applicable
    \item[] {\bf NCBI:} National Center for Biotechnology Information
    \item[] {\bf NLP:} Natural Language Processing
    \item[] {\bf OMIM:} Online Mendelian Inheritance in Man
    \item[] {\bf Prec:} Precision
    \item[] {\bf RAG:} Retrieval Augmented Generation
    \item[] {\bf Rec:} Recall
    \item[] {\bf SemEval:} Semantic Evaluation
    \item[] {\bf SOTA} State-Of-The-Art
    \item[] {\bf Supp.} Supplemental data augmentation strategy
    \item[] {\bf UMLS} Unified Medical Language System
\end{itemize}

\section*{Acknowledgements}
We would like to thank UAB Research Computing for use of their HPC infrastructure. We would like to thank Dr. Maio Danila for manuscript editing and Shan Chen for his advice and comments.

\section*{Funding}
JO received funding from NIH NIAMS grant P30AR072583,"Building and InnovatinG: Digital heAlth Technology and Analytics". GPU compute was supported by an NVidia GPU Grant Program, "Improving Information Extraction in Biomedical Text". Additional funding was provided by NIH NIA grant R01AG057684, "In Silico Screening of Medications for Slowing Alzheimer's Disease Progression".

\section*{Author Contributions}
KS: Software development for synthetic text generation and evaluation, design, analysis and interpretation of the data, and manuscript draft and revisions, SV: software for analysis of concept distribution, RK: statistical analysis JO: conception, software troubleshooting, analysis and interpretation of the data and manuscript draft and revision. All authors read and approved the manuscript.

\section*{Ethics declarations}
\subsection{Ethic approval and consent to participate}
Not applicable.
\subsection{Consent for publication}
Not applicable.
\subsection{Competing interests}
Not applicable.

\begin{appendices}
\section{Concept Unique Identifier (CUI) Attributes in UMLS}\label{secA1}

Using knowledge graph like UMLS to inject knowledge into entity recognition and normalization tool has value, but as shown in Table \ref{tab:cuiratios}, that value is limited when the knowledge consists of little more than the name of the disease condition. There are 93,448 diseases with no synonyms and no definitions.

\begin{table}[h!]
  \begin{tabular}{lllll}
  & \multicolumn{3}{c}{\textbf{Disease CUIs}} \\ \cmidrule(lr){2-4}
    & \textbf{With} & \textbf{Without} & \textbf{CUI Average} & \textbf{Total} \\ \toprule
 Synonyms & 217252 &  $102129$ & $2.84916$ & 909967 \\ 
 Definitions & $53432$ & $265949$ & $0.21725$ & 69417  \\
  \end{tabular}
  \label{tab:cuiratios}
  \caption{Distribution of Synonyms and Definitions for CUIs in the UMLS 2019AB Disease Semantic Group. There are 319381 Disease Group CUIs. }
\end{table}

\section{OOD NER Results}\label{secB1}


\begin{table}[ht]
\resizebox{\textwidth}{!}{%
\begin{tabular}{lccrrrr}
\textbf{Experiment Type} & \textbf{Labelled}              & \textbf{Augmented}             & \textbf{Prec} & \textbf{Rec} & \textbf{Acc} & \textbf{F1} \\\hline
Baseline          & \xmark & \xmark & 0.7915                                         & 0.8000                                           & 0.9772                                        & 0.7957                                      \\ \hline
\multirow{2}{*}{Ideal}              & \xmark & \cmark & 0.7883                                         & 0.7907                                        & 0.9771                                        & 0.7894                                      \\
             & \cmark & \cmark & \textbf{0.8033}               & 0.7839                                        & 0.9765                                        & 0.7933                                      \\\hline
\multirow{2}{*}{Supplemental}       & \xmark & \cmark & 0.7737                                         & 0.7998                                        & 0.9756                                        & 0.7865                                      \\
      & \cmark & \cmark & \textbf{0.7989}               & \underline{\textbf{0.8149}}              & 0.9772                                        & \underline{\textbf{0.8068}}            \\\hline

\multirow{2}{*}{Naive}              & \xmark & \cmark & 0.7742                                         & 0.7986                                        & 0.9750                                         & 0.7860                                       \\
             & \cmark & \cmark & \textbf{0.7974}               & \underline{\textbf{0.8314}}              & \textbf{0.9779}              & \underline{\textbf{0.8140}} \\\hline   
\multirow{2}{*}{Ablation}              & \xmark & \cmark & 0.7844                                         & \textbf{0.8012}& 0.9761                                        & 0.7927                                       \\
             & \cmark & \cmark & \textbf{0.7972}               & \textbf{0.8174}            & 0.9771             & \textbf{0.8072} \\\hline   
\end{tabular}%
}
\vspace*{0mm}
\caption{SemEval 2015 Task 14 DER Performance, \textbf{bold} is a model that beats the baseline, \underline{underline} denotes statistical significance}
\label{tab:Semeval-Reg-NER}
\end{table}

\begin{table}[htbp]
\resizebox{\textwidth}{!}{%
\begin{tabular}{lccrrrr}
\textbf{Experiment Type} & \textbf{Labelled}              & \textbf{Augmented}             & \textbf{Prec} & \textbf{Rec} & \textbf{Acc} & \textbf{F1} \\\hline
Baseline          & \xmark & \xmark & 0.8229                                         & 0.7578                                          & 0.9785                                        & 0.7888                                      \\ \hline
\multirow{2}{*}{Ideal}              & \xmark & \cmark & \textbf{0.8291}& 0.7370                                        & 0.9766                                        & 0.7802                                      \\
             & \cmark & \cmark & \textbf{0.8347}& \textbf{0.7627}& 0.9782                                        & \textbf{0.7970}\\\hline
\multirow{2}{*}{Supplemental}       & \xmark & \cmark & 0.8050                                         & 0.7316                                        & 0.9747                                        & 0.7664                                      \\
      & \cmark & \cmark & \underline{\textbf{0.8411}}& \underline{\textbf{0.7933}}& 0.9784                                        & \underline{\textbf{0.8164}}\\\hline

\multirow{2}{*}{Naive}              & \xmark & \cmark & 0.7965                                         & 0.7360                                        & 0.9752                                         & 0.7650                                       \\
             & \cmark & \cmark & \underline{\textbf{0.8373}}& \underline{\textbf{0.7911}}& \textbf{0.9788}              & \underline{\textbf{0.8135}}\\\hline   
\multirow{2}{*}{Ablation}              & \xmark & \cmark & 0.7956                                         & 0.7362                                        & 0.9749                                        & 0.7647                                       \\
             & \cmark & \cmark & \underline{\textbf{0.8434}}& \underline{\textbf{0.7943}}& 0.9786             & \underline{\textbf{0.8181}}\\\hline   
\end{tabular}%
}
\vspace*{0mm}
\caption{BC5DR DER Performance, \textbf{bold} is a model that beats the baseline, \underline{underline} denotes statistical significance}
\label{tab:BC5DR-Reg-NER}
\end{table}

\begin{table}[htbp]
\resizebox{\textwidth}{!}{%
\begin{tabular}{lccrrrr}
\textbf{Experiment Type} & \textbf{Labelled}              & \textbf{Augmented}             & \textbf{Prec} & \textbf{Rec} & \textbf{Acc} & \textbf{F1} \\\hline
Baseline          & \xmark & \xmark & 0.8587                                         & 0.8052                                          & 0.9809                                        & 0.8310                                      \\ \hline
\multirow{2}{*}{Ideal}              & \xmark & \cmark & 0.8516                                         & \textbf{0.8165}& \textbf{0.9813}& \textbf{0.8337}\\
             & \cmark & \cmark & 0.8547& 0.7905                                        & 0.9798                                        & 0.8213                                      \\\hline
\multirow{2}{*}{Supplemental}       & \xmark & \cmark & 0.8262                                         & 0.7939                                        & 0.9770                                        & 0.8097                                      \\
      & \cmark & \cmark & \textbf{0.8614}               & \underline{\textbf{0.8334}}& 0.9797                                        & \underline{\textbf{0.8471}}\\\hline

\multirow{2}{*}{Naive}              & \xmark & \cmark & 0.8208                                         & 0.7976                                        & 0.9781                                         & 0.8087                                       \\
             & \cmark & \cmark & \underline{\textbf{0.8698}}& \underline{\textbf{0.8404}}& 0.9803& \underline{\textbf{0.8548}}\\\hline   
\multirow{2}{*}{Ablation}              & \xmark & \cmark & 0.8323                                         & 0.7911                                        & 0.9776                                        & 0.8111                                       \\
             & \cmark & \cmark & \underline{\textbf{0.8738}}& \underline{\textbf{0.8415}}& 0.9809             & \underline{\textbf{0.8573}}\\\hline   
\end{tabular}%
}
\vspace*{0mm}
\caption{NCBI DER Performance, \textbf{bold} is a model that beats the baseline, \underline{underline} denotes statistical significance}
\label{tab:NCBI-Reg-NER}
\end{table}

\begin{table}[htbp]
\resizebox{\textwidth}{!}{%
\begin{tabular}{lccrrr}
\textbf{Experiment Type} & \textbf{Labelled}              & \textbf{Augmented}             & \textbf{OOD Prec} & \textbf{OOD Rec} & \textbf{OOD F1} \\ \midrule
Baseline         & \xmark & \xmark & 0.5127                                             & 0.7863                                            & 0.6190                                           \\\hline
\multirow{2}{*}{Ideal}             & \xmark & \cmark & 0.4908                                             & 0.784                                             & 0.6027                                          \\
            & \cmark & \cmark & 0.4564                                             & \textbf{0.8107}& 0.5829                                          \\\hline
\multirow{2}{*}{Supplemental}      & \xmark & \cmark & 0.4979                                             & 0.7740                                             & 0.6044                                          \\
     & \cmark & \cmark & 0.4892                                             & 0.7748                  & 0.5996                                          \\\hline

\multirow{2}{*}{Naive}            & \xmark & \cmark & \textbf{0.5198}                   & 0.7679                                            & 0.6153                \\
            & \cmark & \cmark & \textbf{0.5354}& 0.7710                                             & \underline{\textbf{0.6316}}      \\\hline
\multirow{2}{*}{Ablation}            & \xmark & \cmark & \textbf{0.5145}                   & \textbf{0.7908}& \textbf{0.6230}\\
            & \cmark & \cmark & 0.4948& \textbf{0.7878}& 0.6076\\\hline
\end{tabular}                                  
}
\vspace*{0mm}
\caption{SemEval 2015 Task 14 DER OOD Performance, \textbf{bold} is a model that beats the baseline, \underline{underline} denotes statistical significance}
\label{tab:Semeval-NER-OOD}
\end{table}

\begin{table}[htbp]
\resizebox{\textwidth}{!}{%
\begin{tabular}{lccrrr}
\textbf{Experiment Type} & \textbf{Labelled}              & \textbf{Augmented}             & \textbf{OOD Prec} & \textbf{OOD Rec} & \textbf{OOD F1} \\ \midrule
Baseline         & \xmark & \xmark & 0.4908                                             & 0.8453                                            & 0.6200                                           \\\hline
\multirow{2}{*}{Ideal}             & \xmark & \cmark & 0.4287                                             & \textbf{0.8675}& 0.5728                                          \\
            & \cmark & \cmark & 0.4617                                             & \textbf{0.8659}& 0.6018                                          \\\hline
\multirow{2}{*}{Supplemental}      & \xmark & \cmark & 0.4235                                             & \textbf{0.8552}& 0.5660                                          \\
     & \cmark & \cmark & 0.4770                                             & \textbf{0.8495}& 0.6105                                          \\\hline

\multirow{2}{*}{Naive}            & \xmark & \cmark & 0.4515& 0.8384                                            & 0.5865                \\
            & \cmark & \cmark & 0.4843& \textbf{0.8498}& 0.6168\\\hline
\multirow{2}{*}{Ablation}            & \xmark & \cmark & 0.4464& 0.8441                                            & 0.5835                \\
            & \cmark & \cmark & 0.4832& \textbf{0.8512}& 0.6163\\\hline
\end{tabular}                                  
}
\vspace*{0mm}
\caption{BC5DR DER OOD Performance, \textbf{bold} is a model that beats the baseline, \underline{underline} denotes statistical significance}
\label{tab:BC5DR-NER-OOD}
\end{table}

\begin{table}[htbp]
\resizebox{\textwidth}{!}{%
\begin{tabular}{lccrrr}
\textbf{Experiment Type} & \textbf{Labelled}              & \textbf{Augmented}             & \textbf{OOD Prec} & \textbf{OOD Rec} & \textbf{OOD F1} \\ \midrule
Baseline         & \xmark & \xmark & 0.6648                                             & 0.8932                                            & 0.7622                                           \\\hline
\multirow{2}{*}{Ideal}             & \xmark & \cmark & \underline{\textbf{0.6891}}& 0.8729                                             & \underline{\textbf{0.7697}}\\
            & \cmark & \cmark & 0.6308                                             & 0.8827& 0.7347                                          \\\hline
\multirow{2}{*}{Supplemental}      & \xmark & \cmark & 0.6119                                             & 0.8614                                             & 0.7151                                          \\
     & \cmark & \cmark & 0.6532                                             & 0.8789                  & 0.7494                                          \\\hline

\multirow{2}{*}{Naive}            & \xmark & \cmark & 0.6348& 0.8550                                            & 0.7274                \\
            & \cmark & \cmark & 0.6614& 0.8773                                             & 0.7539\\\hline
\multirow{2}{*}{Ablation}            & \xmark & \cmark & 0.6185& 0.8740                                            & 0.7236                \\
            & \cmark & \cmark & 0.6633& 0.8910                                             & 0.7604\\\hline
\end{tabular}                                  
}
\vspace*{0mm}
\caption{NCBI DER OOD Performance, \textbf{bold} is a model that beats the baseline, \underline{underline} denotes statistical significance}
\label{tab:NCBI-NER-OOD}
\end{table}

\end{appendices}

\FloatBarrier
\bibliography{OzbornPublicationReferenceLibrary}

\end{document}